\documentclass{isprs}

\usepackage{setspace}
\usepackage{geometry}
\usepackage{epstopdf}
\usepackage[labelsep=period]{caption}
\usepackage[british]{babel} 
\usepackage[hang]{footmisc}

\usepackage[authoryear]{natbib}

\usepackage[backref=page]{hyperref}

\usepackage{amsmath}
\usepackage{mathtools}

\usepackage{tabularx}

\usepackage[dvipsnames]{xcolor}

\usepackage{makecell}

\usepackage{xspace}
\usepackage{siunitx}

\usepackage{graphicx}
\graphicspath{{images/}{images/dataset/}{images/overview/}{images/tracking/}{images/stereoConstraint/}{images/matchingComparison/}{images/quantevaluation/}{images/qualevaluation/}{images/stuttgart00/}{images/stuttgart01/}{images/stuttgart02/}{images/kitti0013/}}

\usepackage{tikz}
\usetikzlibrary{shapes, arrows, positioning, calc, patterns, decorations.pathreplacing, angles, quotes}

\usetikzlibrary{shadows}

\providecommand{\tikzsetnextfilename}[1]{} %

\usepackage{subcaption}

\definecolor{cadmiumgreen}{rgb}{0.0, 0.42, 0.24}
\definecolor{cornellred}{rgb}{0.7, 0.11, 0.11}
\definecolor{cornflowerblue}{rgb}{0.39, 0.58, 0.93}
\definecolor{darkgreen}{RGB}{  0,100, 0} 
\definecolor{firebrick}{RGB}{178, 34,34}
\definecolor{gray}{rgb}{0.80, 0.80, 0.80}
\definecolor{hlcolor}{rgb}{1.00, 1.00, 0.5}

\newcommand{\figref}[1]{Fig.~\ref{#1}}                         %
\newcommand{\tabref}[1]{Table~\ref{#1}}                        %

\newcommand{\secref}[1]{Section~\ref{#1}}

	\makeatletter
	\DeclareRobustCommand\onedot{\futurelet\@let@token\@onedot}
	\def\@onedot{\ifx\@let@token.\else.\null\fi\xspace}                %
	\def\eg{\emph{e.g}\onedot} 			%
	             
	\def\ie{\emph{i.e}\onedot} 			%

	\def\wrt{w.r.t\onedot} 				 %
	                    
	\def\etal{\emph{et al}\@onedot}          %
	                
	\makeatother

\renewcommand{\vec}[1]{\ensuremath{\mathbf{#1}}}                   %

\newcommand*{\tran}{^{\mkern-1.5mu\mathsf{T}}}

\definecolor{myblue}{rgb}{0.0, 0.0, 1.0}

\newcommand\hifi[1]{\textcolor{darkgreen}{{\fontseries{b}\selectfont #1}}}
\newcommand\hise[1]{\textcolor{blue}{{\selectfont #1}}}
\newcommand\hith[1]{\textcolor{red}{{\selectfont #1}}}

\newlength{\subfigureWidth}
\begin{document}

\title{3D Surface Reconstruction from Multi-Date Satellite Images}
\version{}

\author{Sebastian Bullinger, Christoph Bodensteiner, Michael Arens}

\address{Fraunhofer IOSB, Ettlingen, Germany \\ \\ \{sebastian.bullinger, christoph.bodensteiner, michael.arens\}@iosb.fraunhofer.de
}

\icwg{}   %

\abstract{
The reconstruction of accurate three-dimensional environment models is one of the most fundamental goals in the field of photogrammetry. Since satellite images provide suitable properties for obtaining large-scale environment reconstructions, there exist a variety of Stereo Matching based methods to reconstruct point clouds for satellite image pairs. Recently, the first Structure from Motion (SfM) based approach has been proposed, which allows to reconstruct point clouds from multiple satellite images. In this work, we propose an extension of this SfM based pipeline that allows us to reconstruct not only point clouds but watertight meshes including texture information. We provide a detailed description of several steps that are mandatory to exploit state-of-the-art mesh reconstruction algorithms in the context of satellite imagery. This includes a decomposition of finite projective camera calibration matrices, a skew correction of corresponding depth maps and input images as well as the recovery of real-world depth maps from reparameterized depth values. The paper presents an extensive quantitative evaluation on multi-date satellite images demonstrating that the proposed pipeline combined with current meshing algorithms outperforms state-of-the-art point cloud reconstruction algorithms in terms of completeness and median error. We make the source code of our pipeline publicly available.
}

\keywords{Satellite Imagery, 3D Reconstruction, Photogrammetry, Surface Reconstruction, Texturing.}

\maketitle

\sloppy

\section{Introduction}
\label{section_introduction}

\subsection{3D Surface Reconstruction with Satellite Imagery}
The creation of large-scale environment reconstructions is relevant for several application areas such as urban planning or environmental monitoring. While there is a variety of modalities and sensors to compute such models, the usage of observation satellites has recently gained significant attraction because of the progress made in the field of image-based reconstruction. In contrast to pictures of ground-level and airborne devices, satellite images cover huge areas, which allow for reconstructing large-scale environment models with less effort. In contrast to active sensors like Lidar or Radar, image data inherently represents appearance information, which allows image-based reconstruction pipelines to derive not only geometry, but also the corresponding texture information. \\
\emph{Stereo Matching} and \emph{Structure from Motion} (SfM) are currently the predominant methods to derive geometric structures from satellite images \citep{FranchisISPRS2014,ZhangCVPRW2019}. While SfM reconstructs information of multiple images, Stereo Matching is restricted to single image pairs. Thus, SfM based approaches are inherently better suited to process large (unstructured) image sets such as multi-date satellite imagery. Since SfM obtains intrinsic and extrinsic camera parameters during reconstruction, the corresponding results allow to utilize dense and surface reconstruction methods to compute dense point clouds, to triangulate corresponding meshes and to perform mesh texturing. In comparison to point clouds, textured meshes oftentimes create a similar appearance with a lower information redundancy and inherently contain specific properties that simplify the computation of consistent visualizations or intersection tests at different scales.

\begin{table*}[tbh]
	\setlength\extrarowheight{-6pt}
	\renewcommand{\arraystretch}{1}
	\newcommand{\sfmcol}{4.1cm}
	\newcommand{\mvscol}{3.6cm}
	\newcommand{\meshcol}{4.25cm}
	\newcommand{\texturecol}{3.5cm}
	\newcommand{\pipelineoffset}{\vspace*{0.1cm}}
	\newcommand{\pipelineheadline}{SfM: & MVS: & Mesh Reconstruction: & Texturing: \\}
	\setlength{\tabcolsep}{6pt}
	\textbf{Colmap (General Purpose Pipeline)} \\
	\begin{tabular}{p{\sfmcol} p{\mvscol} p{\meshcol} p{\texturecol}} 
		\pipelineheadline 
		\cite{Schoenberger2016sfm} & \cite{Schoenberger2016mvs} &  \cite{Kazhdan2013}, \cite{Schoenberger2016mvs} & -
	\end{tabular} \\
	\pipelineoffset \newline
	\textbf{MVE (General Purpose Pipeline)} \\
	\begin{tabular}{p{\sfmcol} p{\mvscol} p{\meshcol} p{\texturecol}} 
	\pipelineheadline
	\cite{MVE} & \cite{GoeseleICCV2007}, \cite{LangguthECCV2016} & \cite{FuhrmannToG2014}, \cite{Ummenhofer2017} & \cite{WaechterECCV2014} 
	\end{tabular} \\
	\pipelineoffset \newline
	\textbf{OpenMVG \& OpenMVS (General Purpose Pipeline)} \\
	\begin{tabular}{p{\sfmcol} p{\mvscol} p{\meshcol} p{\texturecol}} 
	\pipelineheadline
	\cite{Moulon2012ACCV} & \cite{BarnesToG2009} & \cite{JancosekISRN2014} & \cite{WaechterECCV2014}
	\end{tabular} \\
	\pipelineoffset \newline
	\textbf{Meshroom (General Purpose Pipeline)} \\
	\begin{tabular}{p{\sfmcol} p{\mvscol} p{\meshcol} p{\texturecol}}
	\pipelineheadline
	\cite{Moulon2012ACCV} & \cite{HirschmullerCVPR2005} & \cite{JancosekISRN2014} & \cite{BurtToG1983} \\
	\end{tabular} \\
	\pipelineoffset \newline
	\textbf{Ours (Satellite Specific Pipeline)} \\
	\begin{tabular}{p{\sfmcol} p{\mvscol} p{\meshcol} p{\texturecol}}
	\pipelineheadline
	\cite{ZhangCVPRW2019} + & \cite{ZhangCVPRW2019} + & \cite{Ummenhofer2017} & \cite{WaechterECCV2014} \\
	\cite{Schoenberger2016sfm}  & \cite{Schoenberger2016mvs} + &  & \\
	& \cite{GoeseleICCV2007} & & \\
	\end{tabular}
	\caption{Comparison of our pipeline proposed for satellite reconstruction with state-of-the-art (general purpose) reconstruction pipelines including Colmap \protect\citep{Schoenberger2016sfm}, MVE \protect\citep{MVE}, OpenMVG \& OpenMVS \protect\citep{OpenMVG, OpenMVS2020} and Meshroom \protect\citep{Meshroom}.}
\label{table_sota_rec_pipelines}
\end{table*}

\subsection{Paper Overview and Contribution}
We propose a pipeline that allows to compute textured meshes from multi-date satellite imagery. To apply available surface reconstruction libraries to satellite images, several preprocessing steps are required. This leads to the following main contributions: \\
(1) We leverage PAN-sharpening \citep{Pohl1998IJRS} to combine information of image data captured by different sensors, \ie we compute high-resolution satellite images with color information exploiting the high resolution resolution of panchromatic images and the color information of low-resolution multispectral images - see \secref{section_image_processinig}. Since satellites cover huge areas the corresponding reconstructions can easily exceed available memory capacities of modern workstations. Thus, we extract and align only a specific area of interest of the panchromatic and multispectral imagery using the metadata of the respective satellite image. \\
(2) The projection properties of satellite imaging sensors are typically described by \emph{Rational Polynomial Camera} (RPC) models \citep{TaoPERS2001}. Since the reconstruction operates only on a specific (continuous) part of each satellite image, the corresponding (complex) RPC model can be substituted with a \emph{Finite Projective Camera} (FPC) model that approximates the RPC model locally. Since current state-of-the-art dense and surface reconstruction libraries do not support the skew parameter present in finite FPC models, we perform a specific adaption of the reconstructed FPC models that allows us to maintain the consistency of the reconstruction results by performing a skew correction of the corresponding images and depth maps. A mathematical derivation that justifies the selected transformation is provided in \secref{section_substitution}. \\
(3) To improve the numerical stability of the depth map computation, we follow the MVS step proposed by \cite{ZhangCVPRW2019}. Thus, the converted (and undistorted) depth maps are defined \wrt to a reparameterized space. Since \cite{ZhangCVPRW2019} do not provide information on how to obtain the actual depth values, we provide a concise derivation in \secref{section_depth_map_recovery} that demonstrates how the inverse projection matrix allows to recover the actual depth values. \\
(4) Following the steps above we are able to utilize current state-of-the-art dense and surface reconstructions algorithms to reconstruct textured meshes from multi-date satellite imagery - see \tabref{table_sota_rec_pipelines} for the main components of the proposed pipeline. We demonstrate in \secref{section_experiments} the validity of our approach by performing a comprehensive quantitative evaluation, since the computed meshes exceed (in terms of completeness and median error) dense point clouds reconstructed with previously presented methods. \\
(5) To foster future research we make the source code of our pipeline publicly available\footnote{\label{source_code}Source code is available at \url{https://github.com/SBCV/SatelliteSurfaceReconstruction}}.

\subsection{Related Work}

In current literature of satellite image reconstructions there are two types of approaches: Stereo Matching and SfM. While the majority of methods such as \citep{KuschkISPRSA2013}, S2P \citep{FranchisISPRS2014} and ASP \citep{BeyerESS2018} rely on Stereo Matching to compute 3D point clouds from satellite images, VisSat \citep{ZhangCVPRW2019} recently presented an SfM based approach to tackle this problem. \\
While SfM allows to compute a three-dimensional scene representation reflecting the information of multiple images, Stereo Matching is limited to reconstruct point clouds for single image pairs. To apply Stereo Matching in the context of multi-date satellite imagery requires additional post processing steps to fuse the individual point clouds into a single scene representation - see \cite{FaccioloCVPRW2017}. Because of the number of image pairs increase quadratically with the number of input images, Stereo Matching based methods do not scale well for large image sets. SfM tackles the quadratic increase by exploiting specific algorithms and data structures such as scene graphs to accelerate the computation. Thus, SfM requires substantially less computation time than Stereo Matching to reconstruct multi-date satellite images - a corresponding evaluation is provided by \cite{ZhangCVPRW2019}. \\ %
In addition, the output of SfM allows to directly leverage dense and surface reconstruction algorithms for dense point cloud computation, meshing and texturing. Current state-of-the-art dense reconstruction pipelines have been presented in Colmap \citep{Schoenberger2016mvs}, Meshroom \citep{Meshroom}, MVE \citep{MVE} and OpenMVS \citep{OpenMVS2020}, which build on popular algorithms for dense reconstruction \citep{Schoenberger2016mvs,GoeseleICCV2007,LangguthECCV2016,BarnesToG2009}, meshing \citep{Labatut2009,Kazhdan2013,FuhrmannToG2014,Ummenhofer2017,JancosekCVPR2011} and texturing \citep{BurtToG1983,WaechterECCV2014}. \tabref{table_sota_rec_pipelines} shows a comparison of the different image-based (general purpose) pipelines and our proposed approach for multi-date satellite image reconstruction. \\

\section{Satellite Reconstruction Pipeline}
\label{section_reconstruction_pipeline}
\tikzsetnextfilename{pipelineoverview}
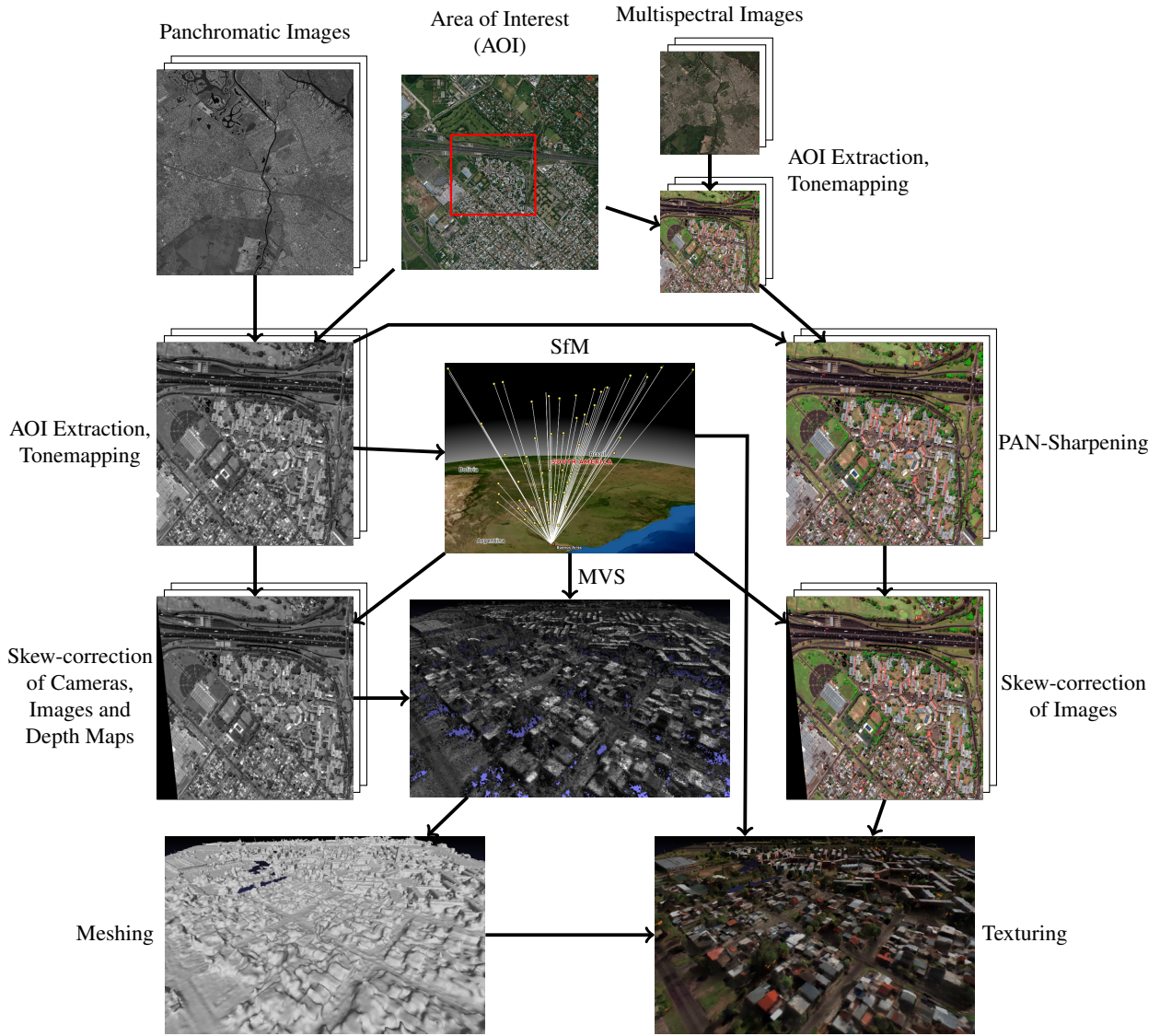
\begin{figure*}[!htb]

	\centering
	\begin{tikzpicture}[auto]
	
		\newcommand{\figureTextSize}{\small}

		\pgfmathsetlengthmacro{\satImageWidth}{1.1in}
		\pgfmathsetlengthmacro{\satImageWidthHalf}{0.55in}
		\pgfmathsetlengthmacro{\SfMWidth}{1.4in}
		\pgfmathsetlengthmacro{\recWidth}{1.8in}
		
		\pgfmathsetmacro{\colDistance}{4.5}
		\pgfmathsetmacro{\rowDistance}{-1.66}
		
		\coordinate (fullRow) at (0, \rowDistance);
		\coordinate (fullCol) at (\colDistance, 0);
		
		\coordinate (halfRow) at (0,0.5 * \rowDistance);
		\coordinate (halfCol) at (0.5 * \colDistance, 0);
		
		\coordinate (diagLineOffsetLD) at (-0.5,-0.25);
		\coordinate (diagLineOffsetRD) at (-0.5,0.25);
		
		\pgfmathsetmacro{\firstcol}{0}
		\pgfmathsetmacro{\firstrow}{0}
		
		\pgfmathsetmacro{\secondcol}{\colDistance}
		\pgfmathsetmacro{\secondrow}{\rowDistance + - 2.25}
		
		\pgfmathsetmacro{\thirdcol}{\secondcol + \colDistance}
		\pgfmathsetmacro{\thirdrow}{\secondrow + \rowDistance - 2}
	
		\pgfmathsetmacro{\fourthcol}{\thirdcol + \colDistance}
		\pgfmathsetmacro{\fourthrow}{\thirdrow + \rowDistance - 1.75}

		\tikzset{imageStack/.style={
				draw=black, 
				double copy shadow={rotate=60, shadow xshift=0.1cm, shadow yshift=0.1cm}, 
				fill=white, 
				inner sep=0}}
			
		\tikzset{reconstructionResult/.style={inner sep=0}}

		\coordinate (descr_offset) at (0,1);
	  
 		\node[imageStack] (PAN) at (\firstcol,\firstrow) { 
	  	\includegraphics[ width = \satImageWidth]{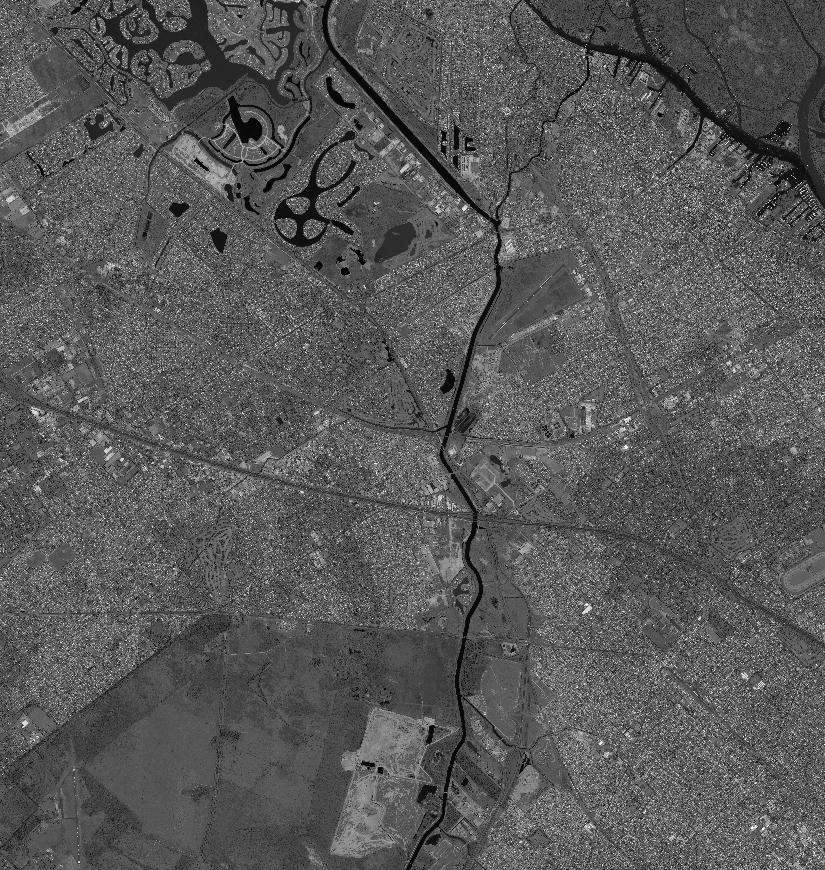}};
  	
  		\node (AOI) at ($(\secondcol, \firstrow) + (-1,0)$){ 
  		\includegraphics[ width = \satImageWidth]{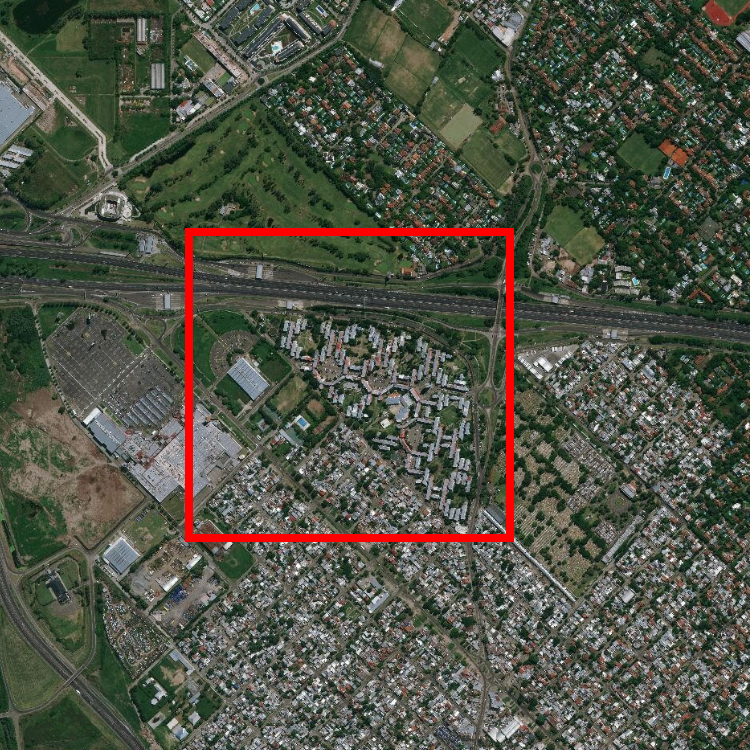}};
		
 		\node[imageStack] (MSI) at ($(\thirdcol,\firstrow) + (-2.5, 1)$) { 
		\includegraphics[ width = \satImageWidthHalf]{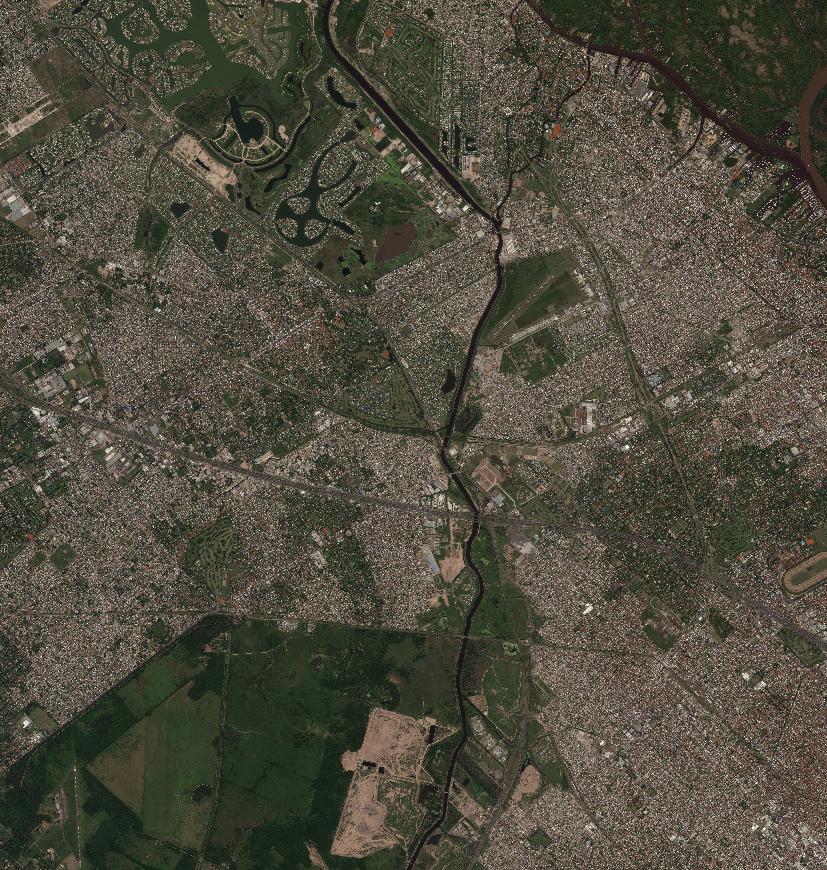}};

		\node[imageStack] (MSIcropped) at ($(\thirdcol, \firstrow) + (-2.5, -1)$){ 
			\includegraphics[ width = \satImageWidthHalf]{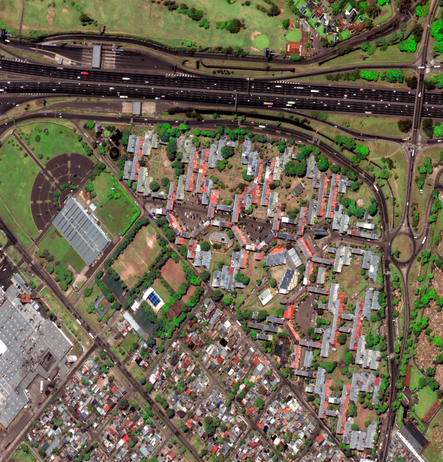}};

		\node[imageStack] (PANcropped) at (\firstcol,\secondrow){ 
			\includegraphics[ width = \satImageWidth]{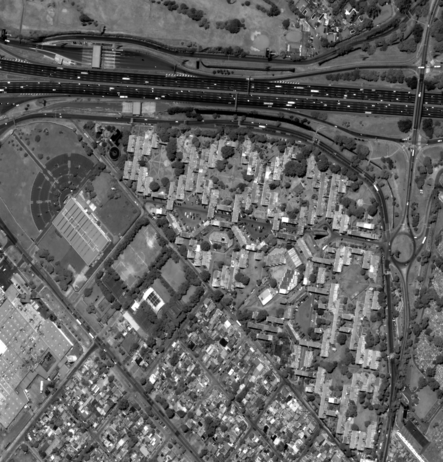}};
		
		\node[imageStack] (PANSharpenedRes) at (\thirdcol, \secondrow){ 
			\includegraphics[ width = \satImageWidth]{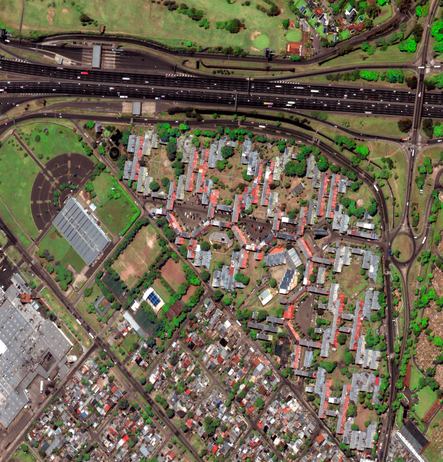}};
		
		\node[reconstructionResult] (sfm) at ($(\secondcol, \secondrow) - (0, 0.2)$){ 
			\includegraphics[ width = \SfMWidth]{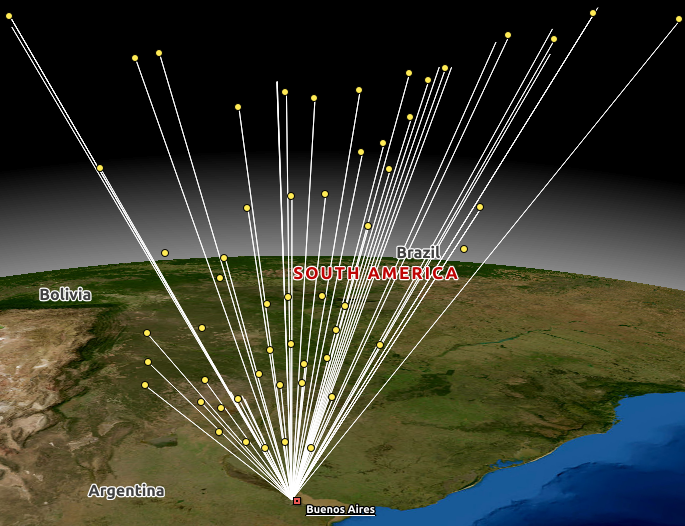}};
	
 		\node[imageStack] (PANskewCorrected) at (\firstcol,\thirdrow) { 
			\includegraphics[ width = \satImageWidth]{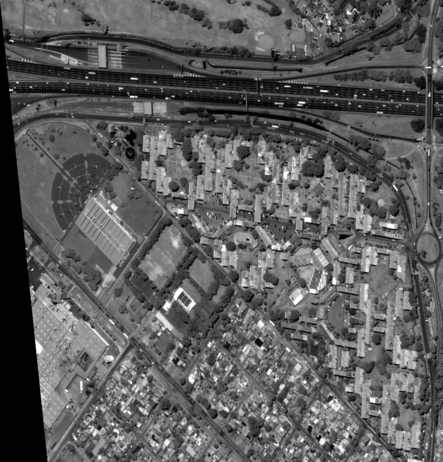}};
		
 		\node[reconstructionResult] (mvs) at (\secondcol,\thirdrow) { 
			\includegraphics[ width = \recWidth]{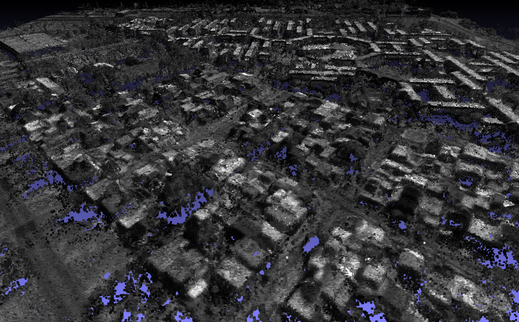}};
		
 		\node[imageStack] (MSIskewCorrected) at (\thirdcol,\thirdrow) { 
			\includegraphics[ width = \satImageWidth]{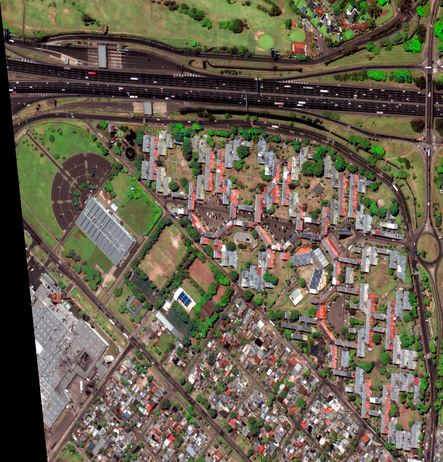}};

		\node[reconstructionResult] (mesh) at ($(\firstcol, \fourthrow) + (1,0)$){ 
			\includegraphics[ width = \recWidth]{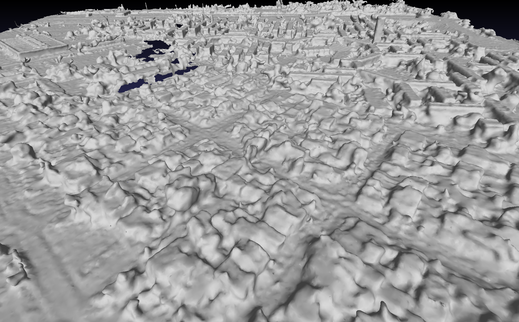}};
		
		\node[reconstructionResult] (texturedMesh) at ($(\thirdcol, \fourthrow) - (1,0)$){ 
			\includegraphics[ width = \recWidth]{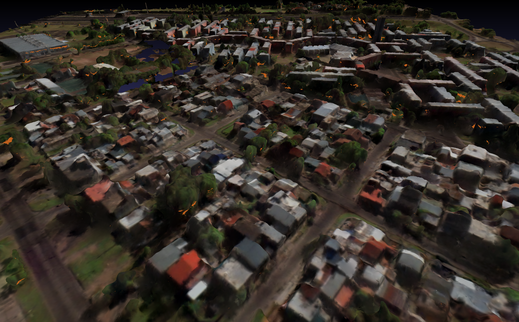}};

	  	\tikzset{arrowStyle/.style={->, line width=0.5mm}}
		\draw[arrowStyle] (AOI) -- (PANcropped.60);
		\draw[arrowStyle] (AOI) -- (MSIcropped);

		\draw[arrowStyle] (PAN.south) -- (PANcropped.north);
		\draw[arrowStyle] (MSI) -- (MSIcropped);
		
		\draw[arrowStyle] (PANcropped.north east) -- ($(PANcropped.north east) - (diagLineOffsetLD)$) -- ($(PANSharpenedRes.north west) + (diagLineOffsetRD)$) -- (PANSharpenedRes.north west);
		\draw[arrowStyle] (MSI) -- (MSIcropped);
		
		\draw[arrowStyle] (PANcropped) -- (sfm);
		\draw[arrowStyle] (PANcropped.south) -- (PANskewCorrected.north);
		
		\draw[arrowStyle] (sfm) -- (PANskewCorrected);
		\draw[arrowStyle] (sfm) -- (mvs);
		\draw[arrowStyle] (sfm) -- (MSIskewCorrected);
		\draw[arrowStyle] ($(sfm.10)$) -- ($(sfm.10) + (0.75,0)$) -- (texturedMesh.125);
				
		\draw[arrowStyle] (PANSharpenedRes.south) -- (MSIskewCorrected.north);
		
		\draw[arrowStyle] (PANskewCorrected) -- (mvs);
		\draw[arrowStyle] (mvs) -- (mesh);
		\draw[arrowStyle] (MSIcropped) -- (PANSharpenedRes.120);
		\draw[arrowStyle] (MSIskewCorrected.south) -- (texturedMesh.60);
		\draw[arrowStyle] (mesh) -- (texturedMesh);

		\tikzset{descrStyle/.style={align=left, font=\figureTextSize}} %

		\node [descrStyle] at ($(PAN)+(0, 2)$) {Panchromatic Images};
		\node [descrStyle, align=center] at ($(AOI)+(0, 2) $) {Area of Interest \\ (AOI)};
		\node [descrStyle] at ($(MSI)+(0, 1.25) $) {Multispectral Images};
		
		\node [descrStyle, align=center] at ($(PANcropped)+(-2.5, 0)$) {AOI Extraction, \\ Tonemapping};
		\node [descrStyle, right] at ($(MSIcropped)+(1,1)$) {AOI Extraction, \\ Tonemapping};
	
		\node [descrStyle, align=center] at ($(PANSharpenedRes)+(2.7, 0)$) {PAN-Sharpening};
		
		\node [descrStyle, align=center] at ($(PANskewCorrected)+(-2.5, 0)$) {Skew-correction \\ of Cameras, \\ Images and \\ Depth Maps};
		\node [descrStyle, align=center] at ($(MSIskewCorrected)+(2.7, 0)$) {Skew-correction \\ of  Images};
		
		\node [descrStyle] at ($(sfm)+(0, 1.6)$) {SfM};
		\node [descrStyle, right] at ($(mvs)+(0, 1.75)$) {MVS};

		\node [descrStyle] at ($(mesh)+(-3.0, 0)$) {Meshing};
		\node [descrStyle] at ($(texturedMesh)+(3.0, 0)$) {Texturing};

	  \end{tikzpicture}

	\caption{Pipeline overview including the different components, the corresponding results and their computational dependencies. As input the pipeline receives a set of panchromatic and multispectral images as well as a bounding box defining the area of interest.} 
\label{reconstruction_pipeline}
\end{figure*}

In the following, we describe the proposed pipeline for reconstructing textured surface models of multi-date satellite imagery. The dependencies of the different pipeline components and their relationships are depicted in \figref{reconstruction_pipeline}.\\
Processing multi-date images of observation satellites with current state-of-the-art image-based reconstruction libraries easily exceeds memory capacities of modern workstations. To control memory requirements and processing times our pipeline determines an area of interest (AOI) in a given set of panchromatic and multispectral image pairs using a single bounding box defined in UTM coordinates. The panchromatic band is agnostic to color information, but shows a lower ground sample distance than the color bands captured by the multispectral sensor.\\
Since many state-of-the-art feature descriptors such as RootSIFT \citep{Arandjelovic2012} exploit image gradients instead of plain color information, we utilize panchromatic images to reconstruct the scene geometry, \ie to perform SfM, MVS and meshing computations. \\
The SfM component exploits the approach by \cite{ZhangCVPRW2019} to locally approximate the RPC model \citep{TaoPERS2001} of each panchromatic image with an FPC model describing only the projection properties of the area of interest. The FPC model has 11 degrees of freedom - the corresponding calibration matrix is shown in \eqref{calib_mat_fin_proj}. In order to represent the obtained SfM reconstruction with camera models containing no skew parameters (\ie 10 degrees of freedom), we apply an intrinsic parameter decomposition of the registered cameras as well as a skew correction of the corresponding input images and depth maps.  Further details are provided in \secref{section_decomposition} and \secref{section_skew_correction}. This representation allows us to utilize current state-of-the-art dense and surface reconstruction libraries to compute dense point clouds and non-textured meshes.\\
In order to texture the reconstructed surfaces, we leverage PAN-Sharpening \citep{Pohl1998IJRS} to compute high-resolution satellite images with color information exploiting the high resolution resolution of panchromatic images and the color information of low-resolution multispectral images. Analogously to the panchromatic images, we apply the skew correction step to the PAN-Sharpened images to utilize current texturing libraries.\\
We considered several state-of-the-art SfM pipelines to reconstruct satellite images as shown in \tabref{table_sota_rec_pipelines}. The evaluation conducted by \cite{Knapitsch2017} and \cite{StathopoulouISPRS2019} suggest that Colmap \citep{Schoenberger2016sfm} is currently one of the top performing SfM pipelines. Conveniently, there exists already VisSat \citep{ZhangCVPRW2019}, a satellite specific adaption of Colmap, that allows us to perform a sparse reconstruction of multi-date satellite images.\\
We considered several methods \citep{Kazhdan2013,Schoenberger2016mvs,FuhrmannToG2014,Ummenhofer2017,JancosekCVPR2011} to reconstruct a watertight surface of the environment. Our quantitative evaluation in \secref{section_experiments} suggests that the algorithm by \cite{Ummenhofer2017} is a reasonable candidate to perform the surface reconstruction in the context of satellite images.\\ 
As shown in \tabref{table_sota_rec_pipelines} our pipelines uses the method by \cite{WaechterECCV2014} in the texturing step.

\section{Satellite Image Preprocessing}
\label{section_image_processinig}
\subsection{Area of Interest Extraction}

As previously mentioned, earth observation satellites typically use panchromatic and multispectral sensors for image acquisition to capture different resolutions and wavelength spectra. The panchromatic band is agnostic to color information, but shows a lower ground sample distance than the color bands captured by the multispectral sensor. We use the provided metadata of the corresponding multispectral image to extract the blue, green and red color band. Further, we filter all panchromatic and multispectral images with high cloud coverage (\ie $>50 \si{\%}$) to reduce the processing time using the corresponding metadata. \\
To compute the area of interest for each panchromatic or multispectral image, we use the corresponding RPC model to project a geo-registered rectangle (provided by the user) to pixel locations.

\subsection{Tone Mapping}

Since the satellite images are defined as high dynamic range images with long-tailed intensity distributions, a naive scaling results in images with low contrast. To tackle this issue we perform the tone mapping procedure proposed by \cite{ZhangCVPRW2019} for panchromatic images that includes a gamma correction and a filtering of intensity values below the $0.5$ percentile and above the $99.5$ percentile. The gamma correction is defined by $I_{g} = I^{1/2.2}$, where $I_{g}$ denotes the gamma corrected image and $I$ the original satellite image. To extend this approach to multispectral images, we apply the percentile filtering for each channel individually. This is necessary to avoid that a single color band dominates the filtering process, which potentially results in unintended color shifts.

\subsection{PAN-Sharpening}

Since the RPC models are geo-registered with a set of image specific ground control points, we are able to project the area of interest to each panchromatic and multispectral image pair. By extracting the resulting sub-parts, we obtain an aligned image pair that represents the same area. Because of the alignment we are able to exploit PAN-Sharpening \citep{Pohl1998IJRS} to leverage the high resolution of the panchromatic image and the color information of the multispectral image to compute a high-resolution image including color information. \\
Recently published PAN-Sharpening approaches \citep{FuCVPR2019,DengIF2019} evaluate the algorithms only on images with low resolution. By conducting several experiments we observed that processing high-resolution imagery with these methods is infeasible because of their high computational requirements. For instance, the implementation by \cite{FuCVPR2019} required on our workstation more than $30 \si{s}$ to process a $100 \times 100$ multispectral and a $400 \times 400$ panchromatic image.
Instead we utilize a weighted variant \citep{GDAL2020} of the widely used Brovey transform \citep{Pohl1998IJRS}. While the algorithm produces overall good results, it occasionally shows artifacts for individual pixels or small pixel regions - caused by strong reflecting surfaces such as vehicles. Since our pipeline computes the final mesh textures with the algorithm proposed by \cite{WaechterECCV2014}, a photo-consistency check ensures that such artifacts do usually not hamper the final appearance.

\section{Substitution of Reconstruction Results}
\label{section_substitution}
We leverage the approach by \cite{ZhangCVPRW2019} in order to locally approximate each pushbroom camera model with a finite projective camera (FPC) model \citep{Hartley2004Book}. Current state-of-the-art libraries for surface reconstruction and texturing such as MVE \citep{MVE} or OpenMVS \citep{OpenMVS2020} do not support the skew parameter contained in finite projective camera models. We address this issue by substituting the reconstructed FPC models with simpler calibration matrices containing no skew factors. To maintain the consistency of the reconstruction we perform a skew correction of the corresponding images and depth maps. \\ 
In the following, we derive a decomposition of the FPC calibration matrix that justifies the selected camera model substitution and the conducted image and depth map transformation.

\subsection{Decomposition of Finite Projective Camera Models}
\label{section_decomposition}

The calibration matrix of the FPC model is shown in \eqref{calib_mat_fin_proj},
\begin{equation}
\vec{K}_p =
\begin{bmatrix} 
f_{x}    & s  & p_{x} \\ 
0    & f_{y}  & p_{y} \\
0 & 0  & 1
\end{bmatrix}
\label{calib_mat_fin_proj}
\end{equation}
where $f_{x}$ and $f_{y}$ represent the focal length of the camera in terms
of pixel dimensions in the $x$ and $y$ direction. Similarly, $p_{x}$ and $p_{y}$ define the principal point with respect to the two pixel dimensions. The value $s$ reflects the skew of the camera. If $s \neq 0$ this means that the pixels are skewed such that the x- and y-axes are not perpendicular. \\
We decompose the calibration matrix $\vec{K}_p$ of the FPC into a simplified calibration matrix $\vec{K}_{p'}$ and an affine transformation $\vec{T}_{p' \rightarrow p}$ with $\vec{K}_p = \vec{T}_{p' \rightarrow p} \vec{K}_{p'} \Leftrightarrow \vec{T}_{p' \rightarrow p} = \vec{K}_p \vec{K}^{-1}_{p'}$. \\
In general, the transformation $\vec{T}_{p' \rightarrow p} = \vec{K}_p \vec{K}^{-1}_{p'}$ of two FPC models $\vec{K}_p$ and $\vec{K}_{p'}$ is given by \eqref{calib_mat_post_transf}. 
\begin{equation}
\begin{aligned}
\underbrace{\begin{bmatrix} 
f_{x}    & s  & p_{x} \\ 
0    & f_{y}  & p_{y} \\
0 & 0  & 1
\end{bmatrix}}_{\vec{K}_p} 
\underbrace{
\begin{bmatrix} 
\frac{1}{f'_{x}} & - \frac{s'}{f'_{x} f'_{y}}  & \frac{p'_{y} s'}{f'_{x} f'_{y}} - \frac{p'_{x}}{f'_{x}} \\ 
0    & \frac{1}{f'_{y}}  & - \frac{p'_{y}}{f'_{y}} \\
0 & 0  & 1
\end{bmatrix} 
}_{\vec{K}^{-1}_{p'}}
= \\
\underbrace{\begin{bmatrix} 
\frac{f_x}{f'_x}    & -\frac{f_x s'}{f'_x f'_y} + \frac{s}{f'_y} & \frac{f_x p'_y s'}{f'_x f'_y} -  \frac{f_x p'_x}{f'_x} - \frac{p'_y s}{f'_y} + p_x \\ 
0    & \frac{f_y}{f'_y}  & -\frac{f_y p'_y}{f'_y} + p_y\\
0 & 0  & 1
\end{bmatrix}}_{\vec{T}_{p' \rightarrow p}} 
\end{aligned}
\label{calib_mat_post_transf}
\end{equation}
The projection of a point in camera coordinates $\vec{p}_{c}$ onto an homogeneous image point $\widetilde{\vec{p}}_{I}$ may now be reformulated according to \eqref{decomposition_p}.
\begin{equation}
\widetilde{\vec{p}}_{I} = \vec{K}_p \vec{p}_{c} \Leftrightarrow \widetilde{\vec{p}}_{I} = \vec{T}_{p' \rightarrow p} \vec{K}_{p'} \vec{p}_{c} \Leftrightarrow \vec{T}^{-1}_{p' \rightarrow p} \widetilde{\vec{p}}_{I} = \vec{K}_{p'} \vec{p}_{c}.
\label{decomposition_p}
\end{equation}
Equation \eqref{decomposition_p} shows that the original calibration matrix $\vec{K}_p$ can be substituted with an arbitrary finite projective camera model $\vec{K}'_{p}$ by transforming the image projections with $\vec{T}^{-1}_{p' \rightarrow p} \widetilde{\vec{p}}_{I}$.

\subsection{Camera Substitution and Skew Correction}
\label{section_skew_correction}
Many state-of-the-art MVS, surface reconstruction and texturing libraries including the (official) implementations used in our evaluation (see \secref{section_experiments}) support only skew-free camera models $\vec{K}_{s}$, \ie $s'$ must be $0$. The choice for $f'_{x}, f'_{y}, p'_{x}$ and $p'_{y}$ is not restricted. By defining $(f'_{x}, f'_{y}, p'_{x}, p'_{y}, s')$ according to $(f'_{x}, f'_{y}, p'_{x}, p'_{y}, s') \coloneqq (f_{x}, f_{y}, p_{x}, p_{y}, 0)$ we obtain the transformation $\vec{T}_{s \rightarrow p}$ shown in \eqref{calib_mat_post_proc}.
\begin{equation}
\underbrace{
\begin{bmatrix} 
f_{x}    & s  & p_{x} \\ 
0    & f_{y}  & p_{y} \\
0 & 0  & 1
\end{bmatrix}
}_{\vec{K}_p} 
\cdot
\underbrace{
\begin{bmatrix} 
\frac{1}{f_{x}} & 0  & \frac{-p_{x}}{f_{x}} \\ 
0    & \frac{1}{f_{y}}  & \frac{-p_{y}}{f_{y}} \\
0 & 0  & 1
\end{bmatrix}
}_{\vec{K}_{s}^{-1}}  
=
\underbrace{
\begin{bmatrix} 
1 & \frac{s}{f_{y}}  & \frac{-s \cdot p_y}{f_{y}} \\ 
0 & 1 & 0 \\
0 & 0 & 1
\end{bmatrix}
}_{\vec{T}_{s \rightarrow p}}
\label{calib_mat_post_proc}
\end{equation}
The matrix $\vec{T}_{s \rightarrow p}$ in \eqref{calib_mat_post_proc} contains not only a skew correction factor (\ie $\frac{s}{f_{y}}$), but also a translation component (\ie $\frac{-s p_y}{f_{y}}$). Thus, a part of the pixels will be translated outside of the image boundaries and important information will be lost. Instead, we define $\vec{K}_{s}$ according to $(f'_{x}, f'_{y}, p'_{x}, p'_{y}, s') \coloneqq (f_{x}, f_{y}, p_{x} - \frac{s \cdot p_y}{f_y}, p_{y}, 0)$ to obtain a translation-free skew-correcting matrix $\vec{T}_{s \rightarrow p}$ as shown in \eqref{calib_mat_post_proc_1}.
\begin{equation}
\underbrace{
\begin{bmatrix} 
f_{x}    & s  & p_{x} \\ 
0    & f_{y}  & p_{y} \\
0 & 0  & 1
\end{bmatrix}
}_{\vec{K}_p} 
\cdot
\underbrace{ 
\begin{bmatrix} 
\frac{1}{f_x}    & 0  & -\frac{p_x}{f_x} + \frac{s \cdot p_y}{f_x \cdot f_y}  \\ 
0    & \frac{1}{f_y}  & -\frac{p_y}{f_y} \\
0 & 0  & 1
\end{bmatrix}
}_{\vec{K}_{s}^{-1}}  
=
\underbrace{
\begin{bmatrix} 
1 & \frac{s}{f_{y}}  & 0 \\ 
0 & 1 & 0 \\
0 & 0 & 1
\end{bmatrix}
}_{\vec{T}_{s \rightarrow p}}
\label{calib_mat_post_proc_1}
\end{equation}
For each registered image $\vec{I}_p$ (corresponding to $\vec{K}_{p}$) we compute a skew-free image $\vec{I}_{s}$ and a skew-free depth map $\vec{D}_{s}$ (corresponding to $\vec{K}_{s}$) using $\vec{I}_{s} = \vec{T}_{s \rightarrow p}^{-1} \cdot \vec{I}_p$ and $\vec{D}_{s} = \vec{T}_{s \rightarrow p}^{-1} \cdot \vec{D}_p$. While removing the skew factors we applied a cubic interpolation to compensate artifacts caused by the discrete nature of pixel based image representations. See \figref{reconstruction_pipeline} for a comparison of the original and the corresponding skew-free image.

\section{Real Depth Map Recovery}
\label{section_depth_map_recovery}
Because the distances of the registered cameras and the reconstructed point cloud (in the context of satellite images) are extremely large, the values in the corresponding depth maps are compressed to an interval that is far away from the origin. To improve the numerical stability of the depth map reconstruction, it is reasonable to perform the computations in a reparameterized space. We follow the reparameterization proposed by \cite{ZhangCVPRW2019}, which uses plane-plus-parallax \citep{IraniECCV1998}. \\
Since \cite{ZhangCVPRW2019} do not describe how to obtain the real depth map values, we provide a concise explanation in the following. The reparameterization is defined according to \eqref{eq_reparameterization},
\begin{equation}
\begin{aligned}
	\underbrace{\begin{bmatrix} 
	\vec{P}_{11}   & \vec{P}_{12}  & \vec{P}_{13} & \vec{P}_{14} \\ 
	\vec{P}_{21}   & \vec{P}_{22}  & \vec{P}_{23} & \vec{P}_{24} \\ 
	\vec{P}_{31}   & \vec{P}_{32}  & \vec{P}_{33} & \vec{P}_{34} \\ 
	0   		   & 0 			   & \bar{Z} & -\bar{Z}d \\ 
	\end{bmatrix}}_{\vec{P}}
	\cdot
	\begin{bmatrix} 
	x  \\ 
	y  \\ 
	z  \\ 
	1  \\ 
	\end{bmatrix} \\
	=
	\begin{bmatrix} 
	uZ  \\ 
	vZ  \\ 
	Z  \\ 
	\bar{Z}z - \bar{Z}d \\ 
	\end{bmatrix} 
	=
	\begin{bmatrix} 
	uZ  \\ 
	vZ  \\ 
	Z  \\ 
	\frac{(\bar{Z}z - \bar{Z}d) Z}{Z} \\ 
	\end{bmatrix}
	\coloneqq
	\begin{bmatrix} 
	uZ  \\ 
	vZ  \\ 
	Z  \\ 
	m Z \\ 
	\end{bmatrix}
	\simeq
	\begin{bmatrix} 
	u  \\ 
	v  \\ 
	1  \\ 
	m \\ 
	\end{bmatrix}
\end{aligned}
	\label{eq_reparameterization}
\end{equation}
where $[x,y,z,1]\tran$ denotes a point in homogeneous world coordinates and $[u,v,1]\tran$ a point in homogeneous image coordinates. $\vec{P}$ represents the (reparameterized) projection matrix from world to image coordinates, $Z$ defines the conventional depth with $Z = \vec{P}_{31} x + \vec{P}_{32} y + \vec{P}_{33} z + \vec{P}_{34}$, $\bar{Z}$ indicates the average of the conventional depth values of the points computed in the SfM step, $(0, 0, 1, d)$ defines a plane below the scene and $m$ denotes the reparameterized depth with $m = (\bar{Z}z - \bar{Z}d) / Z$. \\
Following this reparameterization the MVS step yields reparameterized depth maps representing vectors of the form $[u,v,m]\tran$.\\
By rewriting \eqref{eq_reparameterization} according to \eqref{eq_relation_m_eq_1_div_z}, it becomes apparent that the actual $Z$ value can be obtained using only the reparameterized depth value $m$ and the 4-th (\ie the last) row of $\vec{P}^{-1}$ - in the following denoted as $(\vec{P}^{-1})_4$.
\begin{equation}
\begin{aligned}
\vec{P}
\begin{bmatrix} 
	x  \\ 
	y  \\ 
	z  \\ 
	1  \\ 
\end{bmatrix}
=
\begin{bmatrix} 
uZ  \\ 
vZ  \\ 
Z  \\ 
m Z \\ 
\end{bmatrix}
\Leftrightarrow 
\vec{P}^{-1}
\begin{bmatrix} 
u \\ 
v \\ 
1\\ 
m \\ 
\end{bmatrix} 
=
\begin{bmatrix} 
x/Z  \\ 
y/Z  \\ 
z/Z  \\ 
1/Z  \\ 
\end{bmatrix} \\
\Rightarrow
(\vec{P}^{-1})_4 
\begin{bmatrix} 
u \\ 
v \\ 
1\\ 
m \\ 
\end{bmatrix} 
= 
1 / Z
\Leftrightarrow 
Z = 
\frac{1}{
	(\vec{P}^{-1})_4
	\begin{bmatrix} 
	u, v, 1, m \\ 
	\end{bmatrix}\tran
}
\end{aligned}
\label{eq_relation_m_eq_1_div_z}
\end{equation}
To increase numerical stability, \cite{ZhangCVPRW2019} normalize $\vec{P}$ and $\vec{P}^{-1}$. Let $n_{\vec{P}^{-1}}$ denote the normalization factor of $\vec{P}^{-1}$. We obtain the real depth values using $n_{\vec{P}^{-1}} \cdot Z$. \\
By computing $\vec{K}_{s}$, $\vec{I}_{s}$ (see \secref{section_decomposition}) and $\vec{D}_{s}$ for each registered image, we are able to utilize established libraries for MVS and texturing in the context of satellite images such as the algorithms proposed by \cite{MVE} and \cite{OpenMVS2020}.

\section{Experiments and Evaluation}
\label{section_experiments}

For our experiments we use multi-date satellite imagery of the IARPA MVS3DM dataset \citep{BoschAIPRW2016}. The ground sample distance of the panchromatic and the multispectral images is approximately \SI{0.3}{m} and \SI{1.3}{m}.

\subsection{Quantitative Evaluation}
\label{sec_quantitative_evaluation}

\begin{table*}[tbh]
	\centering
	\setlength{\tabcolsep}{5pt}
	\begin{tabular}{c | c c c | c c c c c c } 
		Pipeline & Depth Map & Meshing & Point &  \multicolumn{2}{c}{Site 1} & \multicolumn{2}{c}{Site 2} & \multicolumn{2}{c}{Site 3} \\

		& Fusion & Algorithm &  Sampling &  CP (\%) & ME (m) & CP (\%) & ME (m) & CP (\%) & ME (m) \\
		
		\hline
		
		Proposed & VisSat & Poisson & Poisson Disk & 72.1 & \hise{0.307} & 66.5 & \hifi{0.411} & 60.9 & \hifi{0.324} \\
		
		Proposed & VisSat & Colmap & Poisson Disk & \hise{73.7} & \hifi{0.305} & \hise{67.9} & \hith{0.433} & \hith{66.8} & \hise{0.428} \\
		
		Proposed & VisSat & OpenMVS & Poisson Disk & \hifi{73.8} & 0.317 & \hise{67.9} & 0.446 & \hise{67.0} & 0.446 \\
		
		Proposed & MVE & FSSR & Poisson Disk & 71.1 & 0.329 & 67.0 & 0.449 & 64.6 & 0.468 \\
		
		Proposed & MVE & GDMR & Poisson Disk & \hith{73.4} & \hise{0.307} & \hifi{68.0} & \hise{0.432} & \hifi{67.4} & \hise{0.428} \\

	\end{tabular}
	\caption{Quantitative evaluation of the proposed pipeline with different state-of-the-art surface reconstruction algorithms. The surface points used to compute the metrics are extracted with poisson disk sampling. The evaluated surface reconstruction methods include Poisson \protect\citep{Kazhdan2013}, Colmap \protect\citep{Schoenberger2016mvs,Labatut2009}, OpenMVS \protect\citep{OpenMVS2020,JancosekISRN2014}, FSSR \protect\citep{FuhrmannToG2014} and GDMR \protect\citep{Ummenhofer2017}. FSSR and GDMR require scale information for each point in the fused point cloud to triangulate the corresponding mesh. In this case we substitute the depth map fusion approach of VisSat \protect\citep{ZhangCVPRW2019} with the equivalent processing step of MVE \protect\citep{GoeseleICCV2007}. The metrics completeness (CP) $\uparrow$ and median error (ME) $\downarrow$ follow the definition by \protect\cite{BoschAIPRW2016}. The top three results are highlighted in green (first), blue (second) and red (third).}
\label{table_quant_eval_mesh_triangulation_pd_hf}
\end{table*}

\begin{table*}[tbh]
	\centering
	\setlength{\tabcolsep}{5pt}
	\begin{tabular}{c | c c c | c c c c c c } 
		Pipeline & Depth Map & Meshing & Point &  \multicolumn{2}{c}{Site 1} & \multicolumn{2}{c}{Site 2} & \multicolumn{2}{c}{Site 3} \\

		& Fusion & Algorithm & Sampling & CP (\%) & ME (m) & CP (\%) & ME (m) & CP (\%) & ME (m) \\
		
		\hline
		
		\cite{ZhangCVPRW2019} & VisSat &  - & Vertex  &  71.7 & 0.340 & 66.8 & 0.460 & 63.2 & 0.413 \\
		
		Proposed & MVE &  - & Vertex & 63.3 & 0.699 & 55.9 & 0.884 & 55.1 & 0.856 \\
		
		Proposed & VisSat & Poisson & Vertex & \hifi{73.4} & \hise{0.305}  & 66.7 & \hifi{0.408}  & 61.1 & \hifi{0.323} \\
		
		Proposed & VisSat & Colmap & Vertex & \hise{73.3} & \hifi{0.284} & \hise{67.7} & \hifi{0.408} & \hith{63.4} & \hise{0.360}  \\
		
		Proposed & VisSat & OpenMVS & Vertex & 71.9 & \hith{0.307} & \hith{67.3} & \hith{0.428} & 62.0 & \hith{0.398} \\
		
		Proposed & MVE & FSSR & Vertex & 72.3 & 0.313 & 66.8 & 0.438  & \hise{64.0} & 0.463 \\
		
		Proposed & MVE & GDMR & Vertex & \hith{73.0} & 0.319 & \hifi{68.0} & 0.445 & \hifi{67.0} & 0.423 \\
		
		\hline

	\end{tabular}
	\caption{Quantitative evaluation of the proposed pipeline with different state-of-the-art surface reconstruction algorithms. The reconstructed mesh vertices are used as surface representation to compute the metrics. See \tabref{table_quant_eval_mesh_triangulation_pd_hf} for a description of the different algorithms and metrics. As baseline for our comparison we use the vertices of the fused point clouds computed with VisSat \protect\citep{ZhangCVPRW2019} and MVE \protect\citep{GoeseleICCV2007}. The top three results are highlighted with green (first), blue (second) and red (third).}
\label{table_quant_eval_mesh_triangulation_vertex_hf}
\end{table*}

\begin{table}[tbh]
	\centering
	\begin{tabular}{c | c c } 
		Meshing & Modified  & Value \\
		Algorithm & Parameter & (Default Value) \\
		\hline
		Poisson & - & - \\
		Colmap & max\_proj\_dist & 0 (20) \\
		OpenMVS & min\_point\_distance & 0 (2.5) \\
		& smoothing\_iterations  & 0 (2) \\
		FSSR & - & - \\
		GDMR & - & -
	\end{tabular}
	\caption{Parameter configuration of the meshing algorithms used for the quantitative evaluations.}
	\label{modified_parameters}
\end{table}

\begin{table}[tbh]
	\setlength{\tabcolsep}{5pt}
	\centering
	\begin{tabular}{c c | c c c } 
		Depth Map & Meshing & Site 1 & Site 2 & Site 3 \\
		Fusion & Algorithm & Vertices & Vertices & Vertices \\
		\hline
		VisSat &  - & 6,289k & 1,880k & 1,780k \\
		MVE &  - & 199,264k & 53,872k & 55,618k \\
		VisSat & Poisson &  11,055k & 4,549k & 4,228k \\
		VisSat & Colmap & 4,861k & 1,424k & 1,381k \\
		VisSat & OpenMVS  & 3,964k & 1,210k & 1,078k \\
		MVE & FSSR & 3,850k & 773k & 825k \\
		MVE & GDMR & 3,353k & 859k & 1,675k
	\end{tabular}
	\caption{Number of vertices reconstructed with the different point fusion and mesh triangulation algorithms for each site.}
\label{table_quant_eval_mesh_triangulation_samples}
\end{table}

To compare the proposed pipeline with previously reported results we utilize the benchmark presented by \cite{ZhangCVPRW2019}, which is based upon the IARPA MVS3DM dataset \citep{BoschAIPRW2016}. The corresponding ground truth consists of a set of geo-registered airbourne lidar scans covering different urban areas near San Fernando, Argentina. \cite{ZhangCVPRW2019} follow the evaluation protocol by \cite{BoschAIPRW2016} to perform a metric comparison of the MVS and the ground truth point cloud. To this end, both point clouds are mapped to a predefined geo-registered grid representing a height map - with cell sizes of $0.5 \times 0.5 \si{m^2}$. The value for each grid cell is defined by the maximum height value of all corresponding points. Since the height map comparison is subject to the initial geo-registration, the evaluation protocol refines the alignment before performing the actual evaluation. \cite{BoschAIPRW2016} define two metric scores that represent the \emph{median error} (ME) and the \emph{completeness} (CP) of the reconstructed height map. Concretely, CP denotes the percentage of cells with an error less than $1 \si{m}$. \\
In order to exploit this benchmark for mesh evaluation, we extract a set of points from the mesh surface. Since the reconstructed vertex positions usually show a non-uniform (and texture dependent) distribution, they are only partly suited to represent the full surface geometry. Thus, we utilize poisson disk sampling \citep{Cook1986} to extract a set of points following a random distribution while ensuring that the distance between any pair of samples is larger than a certain predefined threshold. We follow previous practice \citep{ZhangCVPRW2019} and fill small holes with adjacent values. \\
In order to determine a state-of-the-art surface reconstruction approach for multi-date satellite imagery, we quantitatively evaluate the proposed pipeline with several modern meshing algorithms. \tabref{table_quant_eval_mesh_triangulation_pd_hf} reports the corresponding ME and CP metrics based on surface points extracted with poisson disk sampling. Supplementary, \tabref{table_quant_eval_mesh_triangulation_vertex_hf} provides an analysis using only the reconstructed mesh vertices as surface representation. This allows us to compare the reconstructions with previous results presented by \cite{ZhangCVPRW2019}. If viable, we utilize the default mesh reconstruction parameters whenever possible - an overview of (non-)deviating parameters is presented in \tabref{modified_parameters}. \\
Since points extracted from mesh surfaces with poisson disk sampling are more evenly distributed than reconstructed mesh vertices, they provide a better representation of the corresponding surface. Thus, \tabref{table_quant_eval_mesh_triangulation_pd_hf} contains the main results of our quantitative evaluation. By analyzing the corresponding outcome, we observe that our pipeline achieves the best results with GDMR \citep{Ummenhofer2017}. This approach yields not only the highest completeness score, but also state-of-the-art median height errors. This result is very remarkable considering the fact that GDMR relies on the depth map fusion presented in MVE \citep{MVE}, which produces noisy point clouds compared to the fused points obtained with VisSat. \\
In \tabref{table_quant_eval_mesh_triangulation_vertex_hf} we use the mesh vertices to define a point set representing the reconstructed surfaces to conduct a fair comparison of the triangulated meshes and a set of fused point clouds computed with the approach by \cite{ZhangCVPRW2019}. We observe that the reconstructed mesh vertices outperform the fused points for both metrics, which demonstrates the benefits of the proposed pipeline. Note that our approach achieves (again) the highest completeness scores with GDMR. However, the configurations with Poisson \citep{Kazhdan2013} and Colmap \citep{Schoenberger2016mvs} obtain lower median errors. \\
The different rankings in \tabref{table_quant_eval_mesh_triangulation_pd_hf} and \tabref{table_quant_eval_mesh_triangulation_vertex_hf} demonstrate the importance of selecting an appropriate point sampling method. Note that the evaluation in \tabref{table_quant_eval_mesh_triangulation_vertex_hf} is biased by the number of mesh vertices. For instance, higher vertex numbers will most likely affect the resulting completeness scores. \tabref{table_quant_eval_mesh_triangulation_samples} gives an overview of the corresponding mesh vertices. \\
A comparison of \tabref{table_quant_eval_mesh_triangulation_pd_hf} and \tabref{table_quant_eval_mesh_triangulation_vertex_hf} shows that GDMR sacrifices accuracy of vertex positions to obtain faces closer to the actual geometry, which emphasizes the relevance for selecting an appropriate point sampling method. \\
Because of the non-deterministic behavior of SfM we achieved slightly different results in the depth map computation step compared to the original values reported by \cite{ZhangCVPRW2019}.

\subsection{Qualitative Evaluation}

\begin{figure}[t]
	\captionsetup[subfigure]{aboveskip=2pt}
	\centering
	\begin{subfigure}[t]{\subfigureWidth}
		\includegraphics[width=\subfigureWidth]{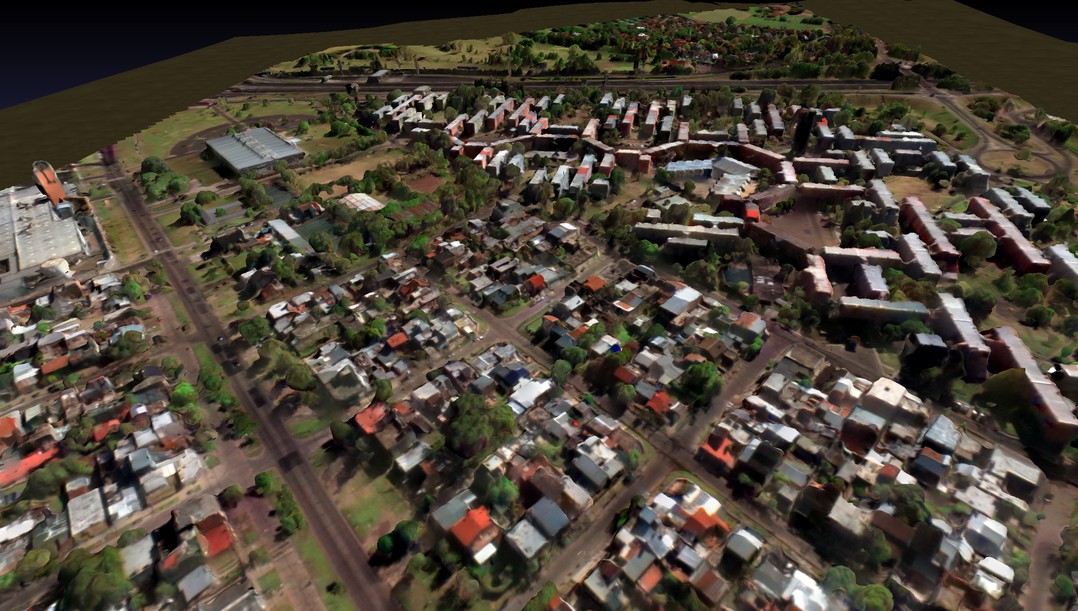}
		\caption{Textured mesh of site 1 reconstructed with GDMR.}
	\end{subfigure}
	\hfill
	\begin{subfigure}[t]{\subfigureWidth}
		\includegraphics[width=\subfigureWidth]{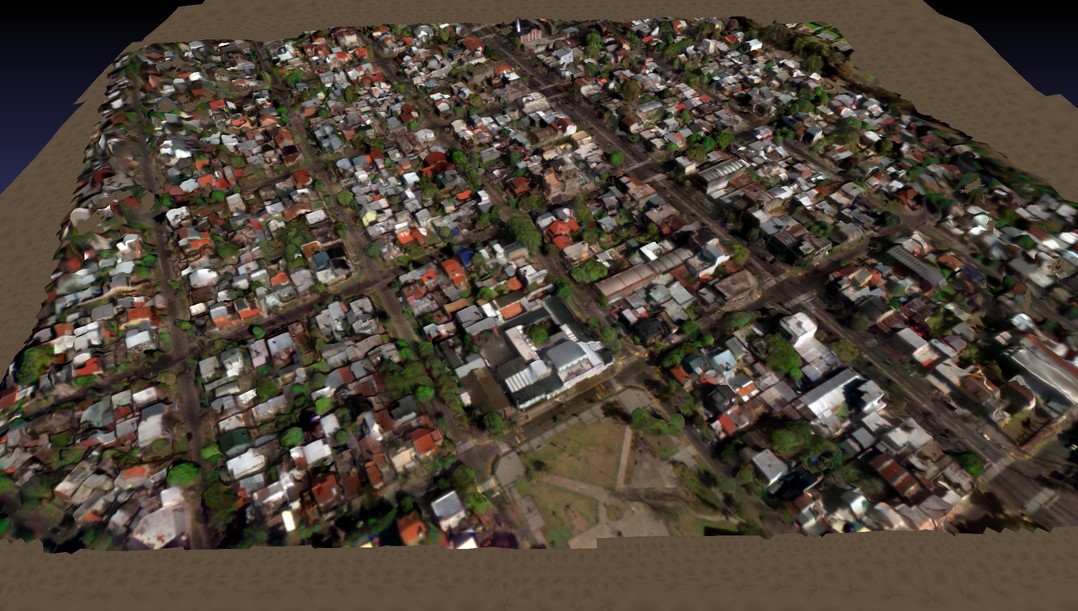}
		\caption{Textured mesh of site 2 reconstructed with GDMR.}
	\end{subfigure}
	\hfill
	\begin{subfigure}[t]{\subfigureWidth}
		\includegraphics[width=\subfigureWidth]{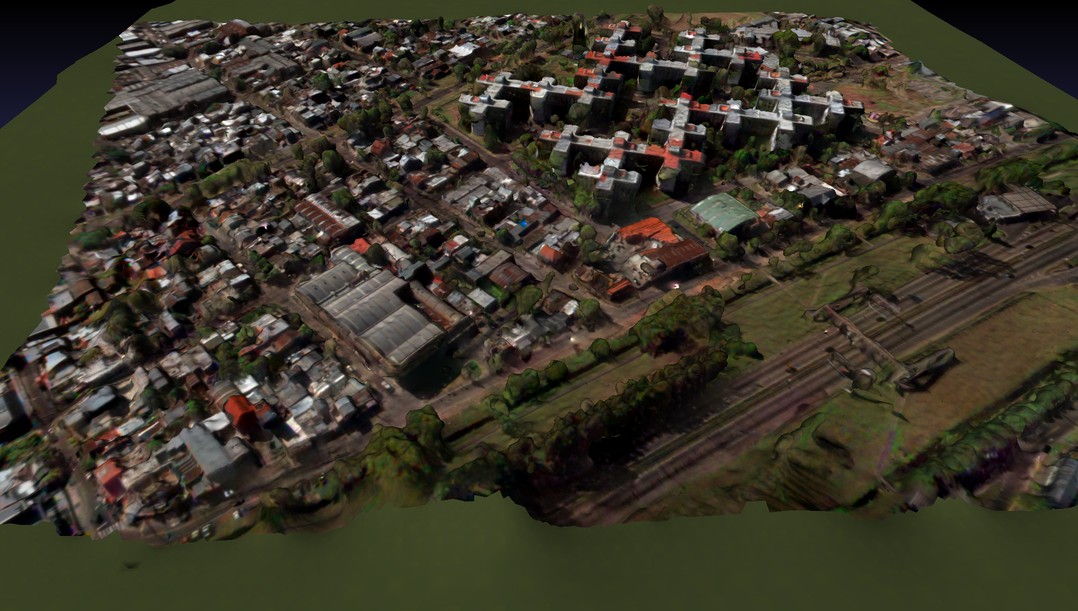}
		\caption{Textured mesh of site 3 reconstructed with GDMR.}
	\end{subfigure}
	\caption{Qualitative evaluation of the proposed pipeline using using GDMR \protect\citep{Ummenhofer2017} for three sites contained in the multi-date satellite image dataset provided by \protect\cite{BoschAIPRW2016}.}
	\label{qualitative_results_textured}
\end{figure}

\begin{figure}[t]
	\captionsetup[subfigure]{aboveskip=2pt}
	\centering
	\begin{subfigure}[t]{\subfigureWidth}
		\includegraphics[width=\subfigureWidth]{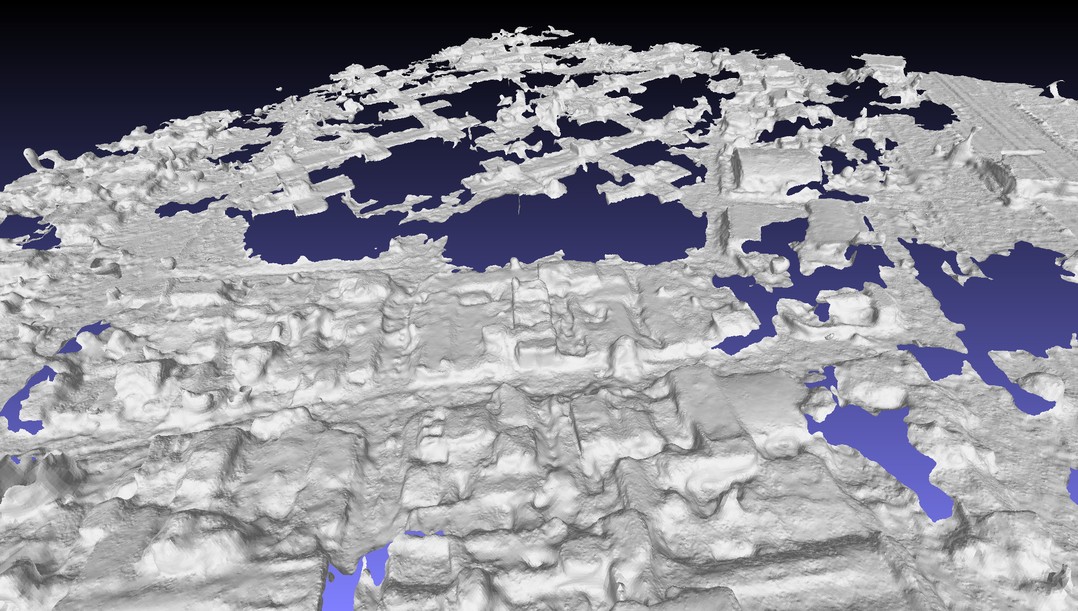}
		\caption{Mesh of site 3 reconstructed with Poisson.}
	\end{subfigure}
	\hfill
	\begin{subfigure}[t]{\subfigureWidth}
		\includegraphics[width=\subfigureWidth]{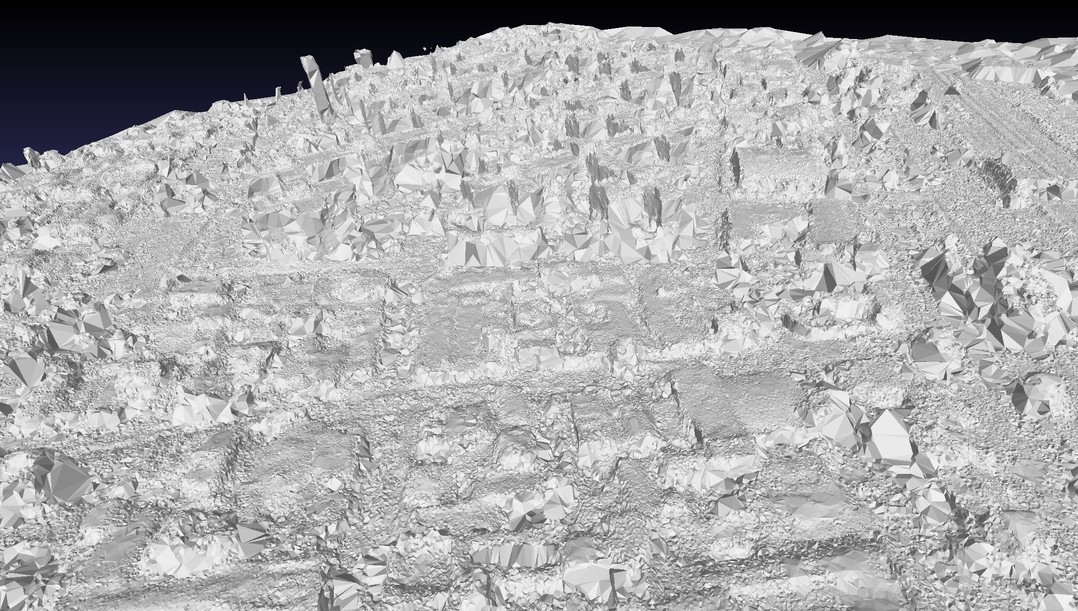}
		\caption{Mesh of site 3 reconstructed with Colmap.}
	\end{subfigure}
	\hfill
	\begin{subfigure}[t]{\subfigureWidth}
		\includegraphics[width=\subfigureWidth]{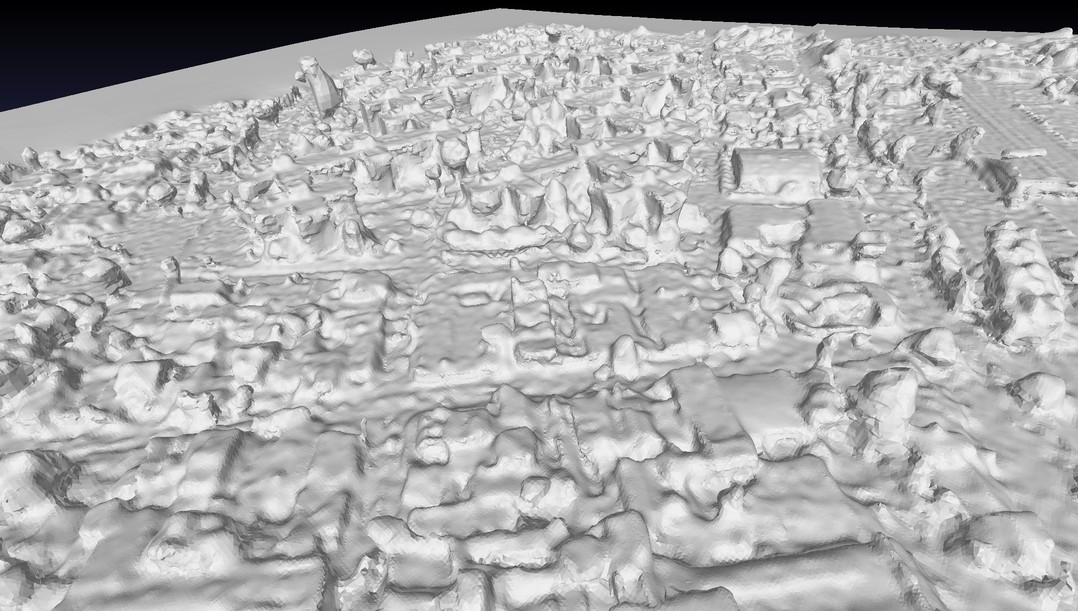}
		\caption{Mesh of site 3 reconstructed with GDMR.}
	\end{subfigure}
	\caption{Qualitative evaluation of our pipeline using multi-date satellite images by \protect\cite{BoschAIPRW2016}. We observe that Poisson \protect\citep{Kazhdan2013}, Colmap \protect\citep{Schoenberger2016mvs} and GDMR \protect\citep{Ummenhofer2017} reconstruct meshes with different shape characteristics.}
	\label{qualitative_results_mesh_characteristics}
\end{figure}

\figref{qualitative_results_textured} shows several reconstructed meshes using the proposed approach with GDMR for meshing, which represents the best configuration of our pipeline. The different reconstructions represent the sites used for the quantitative evaluation. \figref{qualitative_results_mesh_characteristics} visualizes the plain geometry of different meshes obtained with Poisson, Colmap and GDMR. We observe that the reconstructed surfaces exhibit different characteristics (\eg varying levels of completeness, smoothness or mesh granularity), which are consistent to the results reported in \secref{sec_quantitative_evaluation}.

\section{Conclusion}
\label{section_conclusion}
This paper presents an approach to reconstruct textured watertight meshes for multi-date satellite imagery including a detailed description of different pipeline steps and their dependencies. In contrast to previous work, our method leverages current dense and surface reconstruction algorithms to obtain a representation of the scene. We present a decomposition of FPC calibration matrices that allows us to substitute the camera models in the SfM reconstruction with simpler camera models containing no skew factors. To ensure consistent reconstruction results we apply corresponding skew corrections to input images and depth maps. These adjustments enable us to utilize state-of-the-art mesh reconstruction algorithms in the context of satellite images. The paper presents an extensive quantitative evaluation of the pipeline on multi-date satellite images. By sampling points from the reconstructed mesh surfaces we compare our surface reconstructions with point clouds obtained with previously proposed SfM and MVS methods. Our evaluation shows that current state-of-the-art meshing algorithms outperform previous results in terms of completeness and median error. To foster future research we make the source code of the proposed pipeline publicly available.

{
	\begin{spacing}{1.17}
		\normalsize
		\bibliography{satellite_rec} %
	\end{spacing}
}

\end{document}